%% file: paper.tex
\documentclass[lettersize,journal]{IEEEtran}
\usepackage{amsmath,amsfonts}
\usepackage{algorithmic}
\usepackage{algorithm}
\usepackage{array}
\usepackage[caption=false,font=normalsize,labelfont=sf,textfont=sf]{subfig}
\usepackage{textcomp}
\usepackage{stfloats}
\usepackage{url}
\usepackage{verbatim}
\usepackage{graphicx}
\usepackage{cite}
\usepackage{multicol}
\usepackage{color}
\usepackage{multirow}
\usepackage{bm}
\usepackage{bbding}
\usepackage{pifont}
\usepackage{amssymb}
\usepackage{amsmath}
\usepackage{makecell}
\usepackage{booktabs}

\usepackage[colorlinks=true,linkcolor=blue,citecolor=blue]{hyperref}
\usepackage{cleveref}

\crefname{figure}{fig.}{figs.}
\Crefname{figure}{Fig.}{Figs.}

\crefname{table}{table}{tables}
\Crefname{table}{Table}{Tables}

\crefname{equation}{eq.}{eqs.}
\Crefname{equation}{Eq.}{Eqs.}

\begin{document}

\title{PR-DETR: Injecting Position and Relation Prior for Dense \\ Video Captioning}

\author{Yizhe Li,
        Sanping~Zhou,~\IEEEmembership{Member,~IEEE,}
        Zheng~Qin,
        Le~Wang,~\IEEEmembership{Senior Member,~IEEE}
\thanks{This work was supported in part by National Science and Technology Major Project under Grant 2023ZD0121300, National Natural Science Foundation of China under Grants 62088102, U24A20325 and 12326608, and Fundamental Research Funds for the Central Universities under Grant XTR042021005.~({\it Corresponding author: Sanping Zhou, E-mail: spzhou@mail.xjtu.edu.cn.})}
\thanks{Yizhe Li, Sanping Zhou, Zheng Qin and Le Wang are with the National Key Laboratory of Human-Machine Hybrid Augmented Intelligence, National Engineering Research Center for Visual Information and Applications, and Institute of Artificial Intelligence and Robotics, Xi'an Jiaotong University, Xi'an, Shaanxi 710049, China.}}

\markboth{Journal of \LaTeX\ Class Files,~Vol.~14, No.~8, August~2021}%
{Shell \MakeLowercase{\textit{et al.}}: A Sample Article Using IEEEtran.cls for IEEE Journals}


\maketitle

\begin{abstract}
Dense video captioning is a challenging task that aims to localize and caption multiple events in an untrimmed video.
Recent studies mainly follow the transformer-based architecture to jointly perform the two sub-tasks, \emph{i.e.}, event localization and caption generation, in an end-to-end manner. Based on the general philosophy of detection transformer, these methods implicitly learn the event locations and event semantics, which requires a large amount of training data and limits the model's performance in practice. In this paper, we propose a novel dense video captioning framework, named PR-DETR, which injects the explicit position and relation prior into the detection transformer to improve the localization accuracy and caption quality, simultaneously. 
On the one hand, we first generate a set of position-anchored queries to provide the scene-specific position and semantic information about potential events as position prior, which serves as the initial event search regions to eliminate the implausible event proposals. On the other hand, we further design an event relation encoder to explicitly calculate the relationship between event boundaries as relation prior to guide the event interaction to improve the semantic coherence of the captions. Extensive ablation studies are conducted to verify the effectiveness of the position and relation prior. Experimental results also show the competitive performance of our method on ActivityNet Captions and YouCook2 datasets.
\end{abstract}


\section{Introduction}
\label{sec:intro}
\IEEEPARstart{A}{s} a vital branch of video understanding, video captioning~\cite{lin2022swinbert,luo2020univl,seo2022end,pei2019memory,qi2019sports,wang2018reconstruction,cap1} aims to generate a single caption for a short video clip.
To handle the long and untrimmed videos in reality, dense video captioning~\cite{anet,li2018jointly,pdvc,yang2023vid2seq} is proposed to jointly localize multiple events in the video and describe each event with natural language.
Due to its great potential in practical applications, such as video retrieval~\cite{wu2023cap4video,zhang2023multi,ret1} and multimodal  
analysis~\cite{islam2024video,vidchapter, cap2}, dense video captioning has been widely studied since its emergence.
Compared with conventional video captioning, the requirement for accurate multi-event localization and fine-grained captioning makes dense video captioning more difficult in algorithm development, and more practical in engineering deployment.

\begin{figure}[t]
\centering
\includegraphics[width=1.0\linewidth]{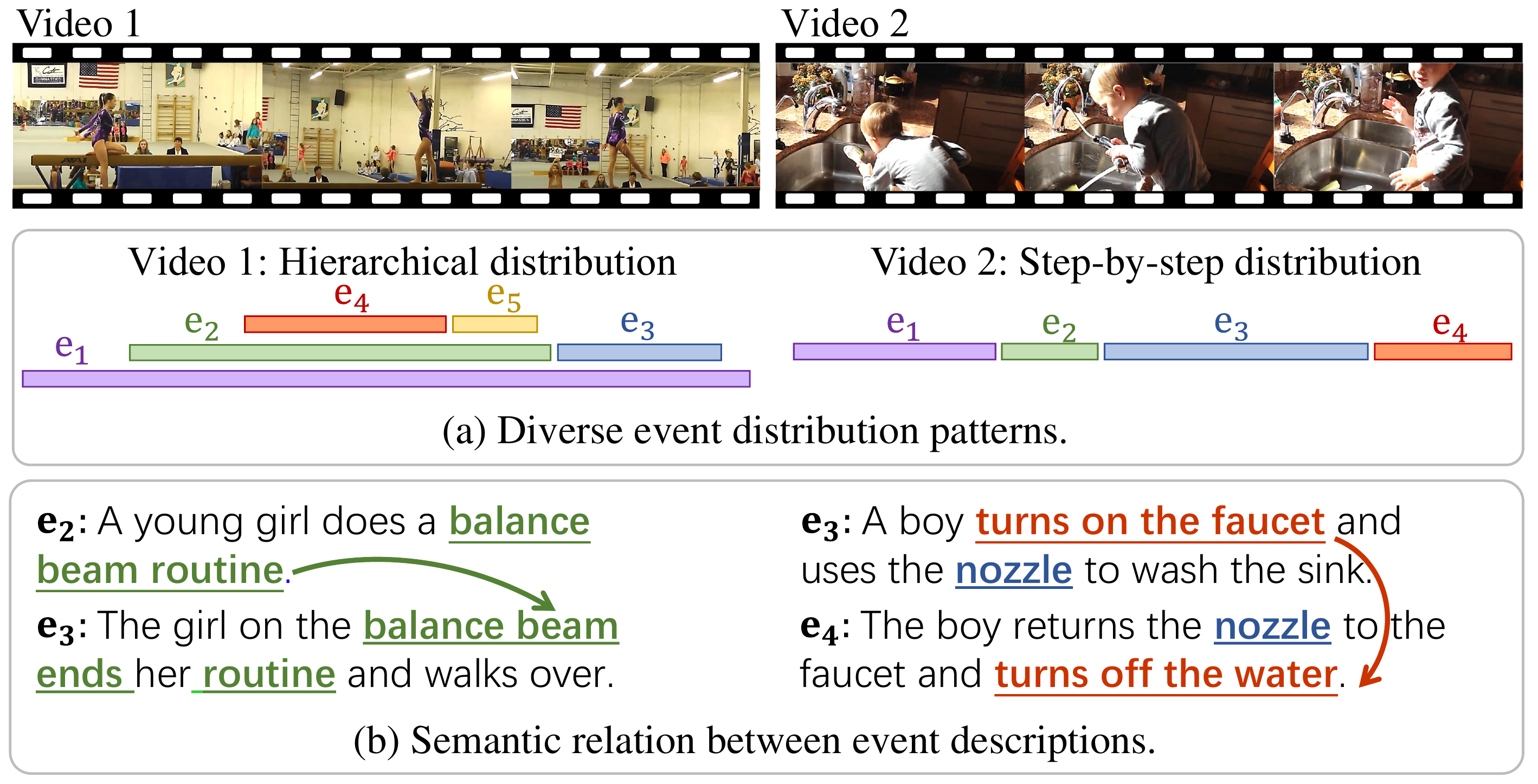}
\caption{Illustration of the event locations and captions in different scenarios within the ActivityNet dataset. (a) The locations of multiple events follow diverse distribution patterns across scenarios. In some scenarios, the events follow a hierarchical distribution and vary widely in length with overlap and inclusion, while in some other scenarios, the events exhibit a step-by-step distribution. (b) There exist logical contexts and co-occurring semantics between event captions within a single video, making it challenging to maintain the coherence of event captions.}
\label{fig:intro}
\end{figure}

Early dense video captioning methods~\cite{iashin2020multi,iashin2020better,yang2018hierarchical} mainly follow a two-stage paradigm, first localizing events in the video, and then generating corresponding captions. In recent years, with the success of DEtection TRansformer~(DETR)~\cite{detr} in object detection, several works~\cite{pdvc,cm2,ppvc,dibs} adopt the DETR-like architecture to perform event localization and caption generation simultaneously.
For example, PDVC~\cite{pdvc} extends DETR into the dense video captioning task, which considers dense video captioning as a set prediction task to predict the locations and captions of multiple events in parallel. CM2~\cite{cm2} enhances the DETR-like model with cross-modal retrieval from external memory. 


Despite the significant advances achieved by these methods, there are still two critical challenges in dense video captioning that limit their further improvement. Firstly, the distribution pattern of events is complex and varies significantly between different scenarios. As shown in~\Cref{fig:intro}~(a), the location of multiple events tends to exhibit a hierarchical distribution in some scenarios, where temporal inclusion and overlap occur between different events. While in some other scenarios, multiple events tend to be relatively short and follow a step-by-step distribution. This diverse distribution of events makes event localization more difficult. Secondly, event descriptions within a single video usually exhibit logical contexts and semantic dependencies, especially for events that are temporally connected. As shown in~\Cref{fig:intro}~(b), event 2 and event 3 in the first scenario include co-occurring semantics such as `balance beam routine'. In the second scenario, events 3 and 4 collectively describe a complete process of washing a sink. This poses a significant challenge for the model to maintain the coherence of the generated captions, constraining the captioning performance of the model.

\begin{figure}[t]
\centering
\includegraphics[width=1\linewidth]{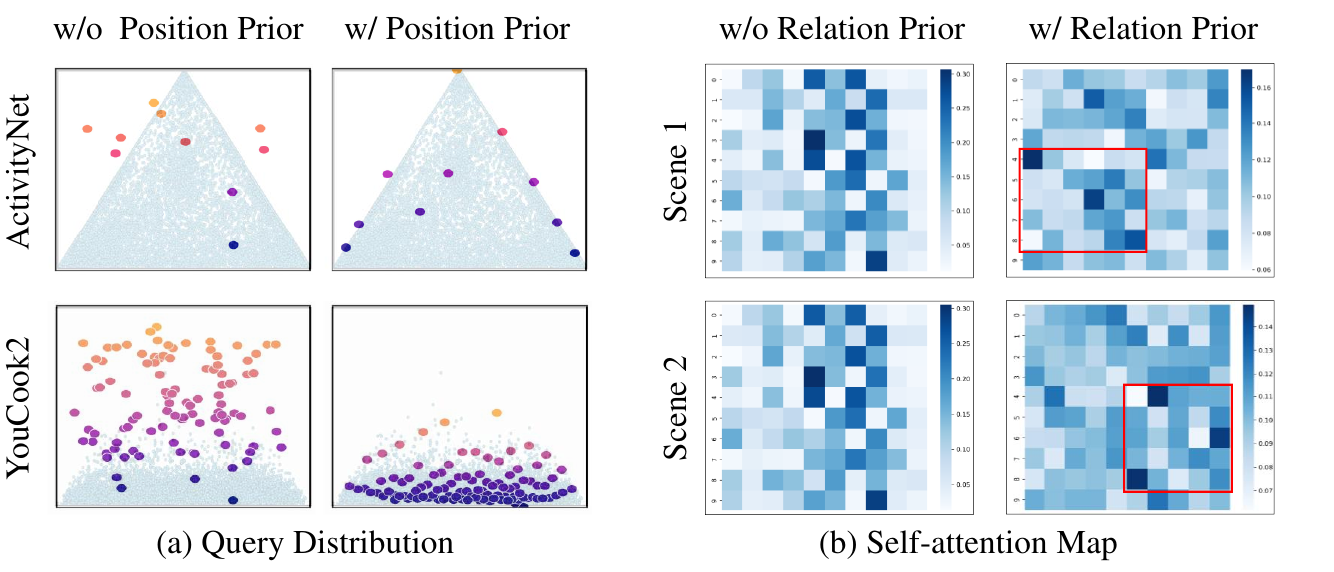}
\caption{(a) Visualization of query distribution with and without position prior. $x$-axis represents the normalized event center coordinates, and the $y$-axis denotes the normalized event duration. (b) Visualization of the self-attention map between event queries with and without relation prior. For different scenarios, previous methods show the same attention map. With the relation prior, the self-attention map better reflects the semantic correlations between potential events.}
\label{fig:fig2}
\end{figure}

To handle the above challenges, previous methods adopt a set of learnable embeddings as event queries to predict the event locations and model the inter-event relations through implicit self-attention between queries. This implicit learning strategy is insufficient for capturing structured event distributions and maintaining consistency across event captions, which may bring ambiguity to the model. As shown in \Cref{fig:fig2} (a), the learned queries are distributed across the entire event space after training, and some of them represent implausible event locations. In the DETR-like architecture, the self-attention map represents the relationship between event queries. As shown in \Cref{fig:fig2} (b), the model has the same attention map for different scenarios, which inadequately reflects the relationship between events. This motivates us to explore the event distribution pattern as the position prior to provide more reliable event search areas. We also model explicit event relations as the relation prior to guide the event interactions in the decoding process, thus improving the localization accuracy and caption quality of the DETR-like dense video captioning models.

In this paper, we propose a novel framework, named PR-DETR, to inject the Position and Relation prior into the DETR-like architecture for the dense video captioning task. For the position prior, we generate a set of position-anchored queries that contain the initial search regions of event location and semantics from the data distribution. 
We first initialize the query with the clustering centers of the ground-truth event location. After {feature aggregation}, the query provides scene-specific search areas of potential events based on the video features, which helps to reduce the ambiguity in the decoding process. For the relation prior, we design an event relation encoder that calculates the explicit pairwise relation between events and then encodes them into an attention mask. This attention mask serves as prior knowledge of event relations, and is integrated into the self-attention operation of the decoder to facilitate better inter-event interaction. With the incorporation of position and relation prior, the location distribution and semantic relations of multiple events in the video are explicitly clarified, which alleviates the training burden of the model, and improves both localization accuracy and caption consistency.


We evaluate our method on two popular dense video captioning datasets: ActivityNet Captions~\cite{anet} and YouCook2~\cite{yc2}.
Our method achieves competitive performance against state-of-the-art methods. Ablation studies also demonstrate the effectiveness of the proposed position and relation prior.
The main contributions of this paper can be summarized as follows:
\begin{itemize}
    \item We design a novel dense video captioning framework, named PR-DETR, which injects position and relation prior to improve the localization accuracy and caption quality, simultaneously.
    \item We design a novel position-anchored query which injects the position prior to provide scene-specific search areas about potential events, improving the event localization accuracy.
    \item We design a novel event relation encoder which injects the relation prior to help capture meaningful event interactions in the decoder, improving the coherence of generated captions.
\end{itemize}

The rest of this paper is organized as follows: We briefly review the related work in Section~\ref{sec:related}. We present the technical details of our proposed method in Section~\ref{sec:method}. Then, extensive experiments and analysis are presented in Section~\ref{sec:experiment}. Finally, we conclude the paper in Section~\ref{sec:conclusion}.

\section{Related Works}
\label{sec:related}

\noindent \textbf{Dense Video Captioning.}
Dense video captioning is a multi-task problem that combines event localization and event captioning to generate informative captions for long and untrimmed videos.
Early approaches~\cite{iashin2020multi,iashin2020better,yang2018hierarchical,wang2018bidirectional,zhou2018end} typically employed a two-stage “localize-then-caption” framework, first localizing events in the video, and then captioning them. For example, Krishna et al.~\cite{anet} first produce event proposals with a proposal module and then generate event captions with an LSTM.
Wang et al.~\cite{wang2018bidirectional} propose a bidirectional proposal method to exploit both past and future contexts for temporal proposal generation. 

However, the separation between event localization and captioning poses a gap in the model, limiting the two tasks to benefit each other.
To solve this problem and enrich inter-task interaction, recent studies~\cite{Deng_2021_CVPR,chadha2020iperceive,chen2021towards,mun2019streamlined,rahman2019watch,shi2019dense} propose to train the two sub-tasks jointly. 
In particular, PDVC~\cite{pdvc} defines dense video captioning as a set prediction task, and proposes an end-to-end framework to conduct event localization and captioning in parallel.
CM2~\cite{cm2}, inspired by the philosophy of Retrieval-Augmented Generation~(RAG)~\cite{rag}, introduces a new dense video captioning method with cross-modal retrieval from external memory.
Vid2Seq~\cite{yang2023vid2seq} leverages unlabeled narrated videos and transcribed speech as multi-modal inputs for pretraining. It predicts a single sequence of discrete tokens that includes caption tokens and special time tokens.
Zhou et al.~\cite{zhou2024streaming} design a streaming model with a memory mechanism and a streaming decoding algorithm to process long videos and live video streams. Instead, our method aims to train a high-quality DVC model from the perspective of prior information without large-scale pretraining.

\noindent \textbf{Transformer-based Architecture.}
Transformer is a deep learning model architecture with attention mechanism for Natural Language Processing (NLP) and other sequence-to-sequence tasks.
Dosovitskiy et al~\cite{vit} first adapt an encoder-only transformer for image recognition, yielding the vision transformer~(ViT), which is widely used in other computer vision tasks~\cite{vit2,detr1,vit4,vit5,vit6,zhou2017point,li2025visual,detr2}.
Carion et al.~\cite{detr} propose DETR, a transformer-based end-to-end object detector without hand-crafted components.
PDVC~\cite{pdvc} extends Deformable Transformer~\cite{zhu2020deformable}, a variant of DETR, into the dense video captioning task, which considers dense video captioning as a set prediction task to predict the locations and captions of multiple events in parallel. 
Based on PDVC, some other approaches~\cite{cm2,ppvc,dibs} are proposed, these methods follow the general philosophy of detection transformer and implicitly learn the position and interaction of multiple events through a set of learnable queries and the attention mechanism.
Instead, our method introduces the task-specific prior into the DETR-like architecture to reduce optimization difficulty and improve model performance.

\noindent \textbf{Event Proposal Generation.}
The previous two-stage methods usually design an event proposal generation module to generate a set of proposals. Some of these methods~\cite{proposal1,proposal2,proposal3} pre-define a vast number of anchors at different scales with regular intervals and utilize an evaluation network to score each anchor. Other works~\cite{proposal4, proposal5, proposal6} predict event boundaries and combine them to form event proposals.
These methods all rely on a trainable network to conduct event proposal generation in a two-stage manner, making a multi-step training strategy necessary. Besides, they contain hand-crafted designs that require laborious tuning and preprocessing.
What's different, our approach generates the position-anchored query which contains 
scene-specific position and semantic information about potential events to serve as proposals, realizing end-to-end proposal generation and training.

\begin{figure}[t]
\centering
\includegraphics[width=1\linewidth]{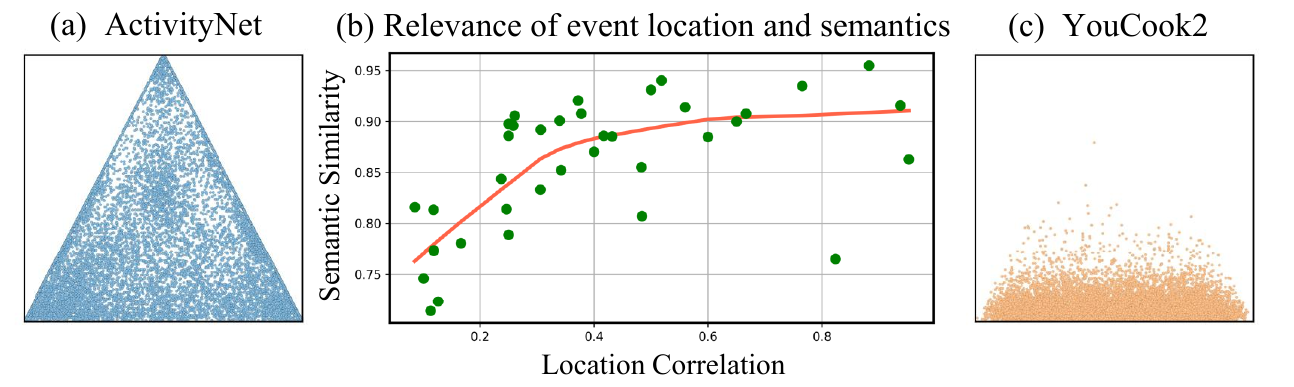}
\caption{Statistical analysis on dense video captioning datasets. (a) The ground-truth event distribution on ActivityNet. (b) Statistical relevance between event location and semantics.
(c) The ground-truth event distribution on YouCook2.
}
\label{fig:stat}
\end{figure}

\section{Method}
\label{sec:method}
\subsection{Statistical Analysis of Event Locations and Semantics}
\label{sec:stat}
In this section, we perform statistical analysis to demonstrate the distribution pattern of event locations and the relevance between event location and semantics. In~\Cref{fig:stat} (a) and (c), we visualize the ground-truth event distributions of two datasets. No matter which dataset, a distinct constraint can be observed in the distribution of event locations, that is, relatively long events tend to occur in the center of the video, which narrows down the search area for events.

To quantify the relevance between event location and semantics, we first design an IOU-style metric to measure the location correlation between the temporal boundaries of two events. 
Given two event boundaries $(t^s_i,t^e_i)$ and $(t^s_j,t^e_j)$ within a video, we compute the Location Correlation~(LC) between them as follows:
\begin{equation}
	\vspace{-3pt}
  \text{LC} = 1 + \frac{\text{min}(t^e_i, t^e_j)-\text{max}(t^s_i, t^s_j)}{\text{max}(t^e_i, t^e_j)-\text{min}(t^s_i, t^s_j)},
\end{equation}
where $t^s_i$ and $t^e_i$ denote the normalized start and end time of the $i$-th event respectively. For the semantic similarity between two event descriptions, we adopt the BERT~\cite{bert} model to extract features of event descriptions and calculate their cosine similarity. \Cref{fig:stat} (b) illustrates the location correlation and semantic similarity of multiple events within a single video. As shown in the figure, there is a positive correlation between event locations and semantics, which means that two temporally adjacent events tend to have more relevant semantics.
{The above statistical results encourage us to explore and leverage the task-specific prior knowledge to help the model training for better localization and captioning.}



\subsection{Overview of PR-DETR}

\begin{figure*}[t]
\centering
\includegraphics[width=\linewidth]{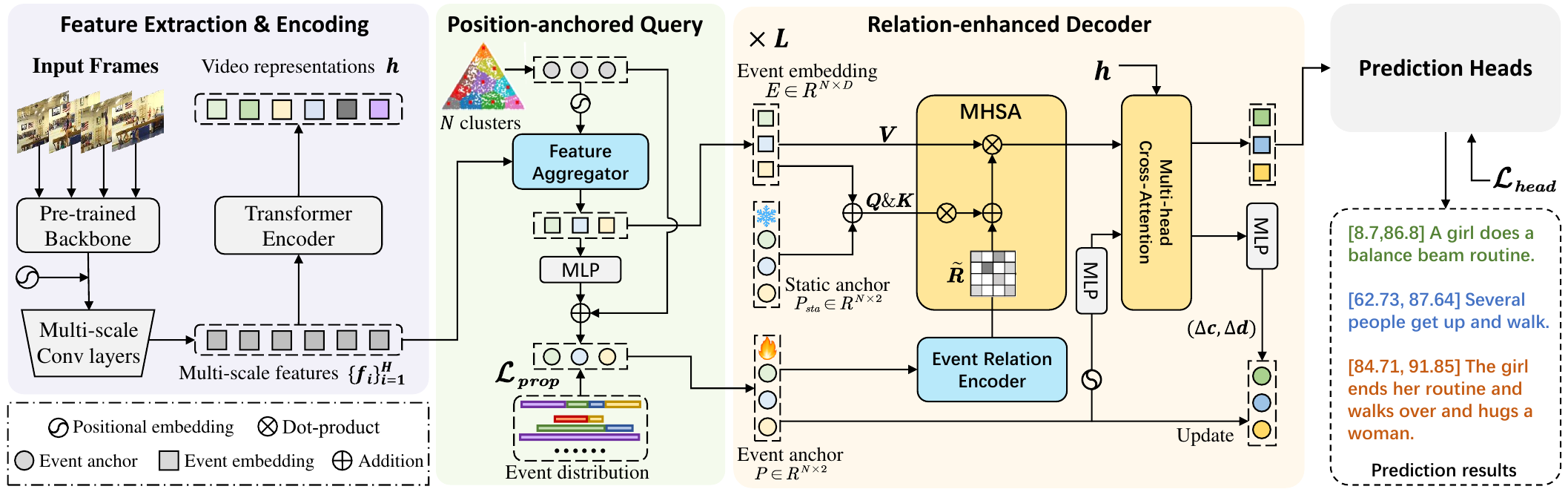}
\caption{Illustration of the proposed PR-DETR. We first obtain the video representations from the input video in feature extraction and encoding. Next, we construct and introduce the position and relation prior through the position-anchored query and relation-enhanced decoder. Finally, the output embedding of the decoder is fed into the prediction heads to produce the temporal boundaries and captions of dense events in the video.
}
\label{fig:pipeline}
\end{figure*}

We illustrate the framework of our PR-DETR in~\Cref{fig:pipeline}. Given an untrimmed video $\mathcal{V}$ as input, our model outputs a set of event locations and corresponding captions \(\{(t^s_n,t^e_n,\mathcal S_n)\}^{N}_{n=1}\), where $N$ denotes the number of events and $\mathcal S_n$ denotes the generated caption for $n$-th event. 
For the input video $\mathcal{V}$, we first perform feature extraction and encoding (Section~\ref{sec:encoding}) to obtain the encoded video representations \(\mathbf{h}\).
Then, we design a set of position-anchored queries to probe the encoded video features $\mathbf{h}$ through the relation-enhanced decoder in dense event decoding (Section~\ref{sec:decoding}), producing the event-level features with rich semantic information. 
After that, we employ several parallel heads to predict the location and caption of events from the event-level features~(Section~\ref{sec:head}).
Finally, we present the training and inference of PR-DETR in Section~\ref{sec:training}.

\subsection{Feature Extraction and Encoding}
\label{sec:encoding}
Given a video sequence $\mathcal{V}$, we first adopt a pre-trained backbone network CLIP~\cite{radford2021learning} to extract the $D$-dimensional frame-level features at 1 FPS. Then, the frame-level features are re-scaled to a fixed length ${V}$ by interpolation for batch processing. To better perceive and predict events of varying durations in the video, we utilize $H$ convolution layers to downsample the frame-level features to different temporal scales, obtaining the multi-scale features $\{\mathbf{f}_i\}^H_{i=1}$, where $\mathbf{f}_i\in  \mathbb{R}^{{V}/{2^{(i-1)}}\times D}$.
Finally, we feed $\{\mathbf{f}_i\}^H_{i=1}$ into the transformer encoder along with their positional embedding~\cite{transformer} to perform the inter-frame feature interaction in different temporal scales. {The transformer encoder is designed with a stack of deformable transformer encoder layers, where each layer consists of a self-attention operation and a feedforward network. Through the encoder, we obtain the encoded video representations $\mathbf{h}$.}






\subsection{Dense Events Decoding}
\label{sec:decoding}
In this section, we elaborate on the decoding process of our method, which includes how we construct the position and relation prior and integrate them into our framework.

\noindent \textbf{Position-anchored Query.}
In the DETR-like methods, a set of event queries is used to probe the encoded features $\mathbf{h}$ for dense event prediction. Previous works initialize the event query with learnable embeddings. The design of the high-dimensional and scene-agnostic query makes the search area ambiguous and training difficult.

To alleviate this problem, we generate a set of position-anchored queries from the data distribution. It consists of two parts: high-dimensional event embedding and 2-dimensional event anchor. The event embedding serves as the carrier of event semantics, and the event anchor represents explicit event locations $\left(\mathbf c, \mathbf d\right)$, where $\mathbf c$ and $\mathbf d$ denote the event center and duration, respectively. We first obtain $N$ centroids from the ground-truth data distribution through the k-means clustering algorithm, where $N$ represents the number of queries. After that, we take the $N$ initial event locations as anchors to aggregate the features of semantically related frames into event-level features through feature aggregation. The obtained event-centric features are used as the event embedding. Then, we refine the initial $N$ clustering boundaries based on the event embedding to get the event anchors. 

Specifically, we flatten the multi-scale features $\{\mathbf{f}_i\}^H_{i=1}$, denoted as $\mathbf{F}$, then we project $N$ clustering centroids into the high-dimensional space through the sinusoidal positional encoding to obtain $N$ event slots $\mathbf{S}\in \mathbb{R}^{N\times D}$. 
In the feature aggregator, we interact the event slots $\mathbf{S}$ with frame-wise features $\mathbf{F}$ for $K$ iterations. In $k$-th iteration, we first transform $\mathbf{S}^k$ and $\mathbf{F}$ as follows:
\begin{equation}
	\begin{aligned}
	\mathbf {\widetilde{S}}^{k}=\Phi \left(\mathtt{LN}(\mathbf S^{k})\right), \quad \mathbf{\widetilde F} =\Phi \left(\mathtt{LN}({\mathbf F})\right),\\
	\end{aligned}
\end{equation}
where $\Phi(\cdot)$ represents a linear layer and $\mathtt {LN}(\cdot)$ denotes layer normalization.
Then, we calculate the attention weight matrix $\mathbf{A}^k_{attn}$ as follows:
\begin{equation}
	\mathbf{A}^k_{attn} = \mathtt{Softmax}\left(\frac{\mathbf{\widetilde S}^{k}{\mathbf{\widetilde F}}^\top}{\sqrt{D}}\right),
\end{equation}
where $D$ is the hidden dimension of the model.
We normalize the attention weight over the slot dimension to ensure the attention weights sum to one for each visual feature, obtaining $\widetilde {\mathbf{A}}^k_{attn}$, where $\widetilde{\mathbf{a}}^k_{m,n}=\frac{\mathbf{a}^k_{m,n}}{\sum_{m=1}^M\mathbf{a}^k_{m,n}}.$ 
The event slot is updated from $\mathbf{S}^{k}$ to $\mathbf{S}^{k+1}$ as follows:

\begin{equation}\label{eq:event}
\begin{aligned}
    &\mathbf U = \mathbf S^{k} + \widetilde{\mathbf{A}}^k_{attn} \Phi (\mathtt{LN}(\mathbf F)),\\
    &\mathbf S^{k+1} = \mathbf U + \mathtt{MLP}\left(\mathtt{LN}(\mathbf U) \right).\\
\end{aligned}
\end{equation}

After $K$ iterations, the event slot $\mathbf{S}^K$ contains preliminary semantic and location information of each potential event in the video. Therefore, we take $\mathbf{S}^K$ as the initial event embedding $\mathbf{E}\in \mathbb{R}^{N\times D}$ and utilize an MLP to 
project $\mathbf{S}^K$ into 2D-temporal offsets. Then we add the $N$ clustering centers with the offsets to obtain the scene-specific event anchor $\mathbf{P}= \{(c_n,d_n)\}^{N}_{n=1} \in \mathbb{R}^{N\times 2}$. The event anchor and event embedding serve as the initial location and semantic search proposal of the potential events in the decoding process, respectively.


 To supervise the generation of position-anchored queries, we introduce the proposal loss $\mathcal L_{prop}$. We first use the Hungarian algorithm~\cite{hungarian} to find the optimal bipartite matching $\delta \in P_n$ between the predicted event anchor $\mathbf P$ and the ground-truth event location $\hat{\mathbf P}$ by minimizing the matching cost as follows:

\begin{equation}
\vspace{-3pt}
	\hat{\delta} = \mathop{\arg \min} \limits_{\delta \in P_n} \mathcal{L}_{giou}(\mathbf{P}_{\delta(i)}, \hat{\mathbf{P}}_{i}),
\end{equation}
where $\mathcal{L}_{giou}$ is the generalized IOU loss function~\cite{giou}.
After obtaining the best matching $\hat{\delta}$ with the lowest matching cost, we use it as a new index of predictions to calculate the proposal loss as follows:

\begin{equation}\label{eq:event}
\begin{aligned}
    \mathcal{L}_{prop}=\sum\limits_i^N\mathcal{L}_{giou}({\mathbf{P}}_{\hat{\sigma}(i)}, \hat{\mathbf{P}}_i).
\end{aligned}
\end{equation}


\noindent \textbf{Relation-enhanced Decoder.}
Although the position-anchored query provides the initial search area of events for the model, the relationship between events should also be considered to generate more coherent and reasonable captions.
Previous methods implicitly learn the interaction between events from training data through the attention mechanism, increasing the optimization difficulty. Instead, we construct and encode the relationship between events as an explicit prior and introduce it into self-attention for better inter-event interactions in the relation-enhanced decoder. 

Specifically, since the event anchor $\mathbf{P}$ contains the distribution pattern of the events, we design static anchors $\mathbf P_\text{sta}=\mathbf{P}$ to maintain the position guidance during decoding. In the relation-enhanced decoder, the static anchor $\mathbf P_\text{sta}$ maintains the position prior without updating, while the event anchor $\mathbf{P}$ is updated layer-by-layer for event localization. To match the input dimension of the self-attention, we project $\mathbf P_\text{sta}$ to $D$-dimensional space through the sinusoidal positional encoding and an MLP as follows:

    \begin{equation}
             \widetilde{\mathbf P}_\text{sta} = \mathtt{MLP}\left(\mathtt{PE}(\mathbf P_\text{sta})\right),
    \end{equation} 
where $\mathtt {PE}(\cdot)$ denotes the sinusoidal position encoding.

Motivated by the observation that temporally connected events exhibit a stronger semantic correlation, we design an overlap-aware distance following IOU-like metrics to indicate whether two event boundaries are temporally connected. The overlap-aware distance of two temporally connected events is 0, whereas overlapping or distant events are computed as close to 1. At $l$-th decoder layer, we first calculate the pairwise overlap-aware distance between the $N$ event anchors $\mathbf P=\{(c_n,d_n)\}^{N}_{n=1}$ to obtain relation matrix $\mathbf R$ as follows:

\begin{equation}
\begin{aligned}
\label{eq:relation}
    {\beta} &= |\text{min}(t^e_i, t^e_j)-\text{max}(t^s_i, t^s_j)|,\\[2mm]
  \bm r(\bm p_i, &\bm p_j) =\left[\log\left(\frac{\beta}{d_i}\!+\!1\right)\!, \log\left(\frac{d_i}{d_j}\right)\!\right],
\end{aligned}
\end{equation}
where $t^s_i$ and $t^e_i$ denote the normalized start and end time of the $i$-th event, $\bm r(\bm p_i,\bm p_j)$ is an element of $\mathbf R$ and represents the relative positional relation between the $i$-th the $j$-th events.
After obtaining the relation matrix $\mathbf R$, we adopt sinusoidal positional encoding to embed each element into a high-dimensional embedding. Then, a 1 $\times$ 1 convolution layer with normalization and activation function is used to perform the linear transformation on it as follows:
\begin{equation}
    \widetilde{\mathbf R} = \mathtt{ReLU}\left(\mathtt{LN}\left(\mathtt{Conv}(\mathtt{PE}\left(\mathbf R)\right)\right)\right) \in \mathbb{R}^{N\times N \times M},
\end{equation}
where $\widetilde{\mathbf R}$ is the relation mask, $M$ corresponds to the number of attention heads. Then, we integrate the relation mask $\widetilde{\mathbf R}$ into self-attention. $\mathbf{Q}$, $\mathbf{K}$, and $\mathbf{V}$ used to perform self-attention are calculated as follows:
\begin{equation}
	\begin{aligned}
	\mathbf{Q}=\mathbf{W}_q (\mathbf E^{l} \oplus                      \widetilde{\mathbf P}_\text{sta}) ,    
	\mathbf{K}=\mathbf{W}_k (\mathbf E^{l} \oplus                      \widetilde{\mathbf P}_\text{sta}),
	\mathbf{V}=\mathbf{W}_v \mathbf E^{l},
	\end{aligned}
\end{equation}
where $\mathbf{W}_q$, $\mathbf{W}_k$, $\mathbf{W}_v$ are linear projection weights and $\oplus$ denotes the element-wise addition operation. Based on $\mathbf Q$, $\mathbf K$ and $\widetilde{\mathbf R}$, we compute the attention weight and obtain the updated event embedding $\widetilde{\mathbf E}$ as follows:
\begin{equation}
        \label{eq:attn}
	\widetilde{\mathbf E} = \mathtt{Softmax}\left(\widetilde{\mathbf R} + \frac{\mathbf{Q}\mathbf{K}^\top}{\sqrt{D}}\right)\mathbf{V} + \mathbf E^{l},
\end{equation}
where $D$ is the hidden dimension of the model.
After self-attention, we take $\widetilde{\mathbf E}$ as the query to interact with the video representations $\mathbf{h}$ through multi-head cross-attention~(MHCA) and a feed-forward network~(FFN), producing the updated event embedding $\mathbf E^{l+1}$.



\subsection{Predction Heads}
Following PDVC, we employ a parallel head structure, which includes a localization head, a captioning head, and an event counter to simultaneously predict $N$ event boundaries, $N$ captions, and event count from the event embedding of the final decoder layer $\mathbf E^L$.

\noindent \textbf{Localization Head.}
The localization head is designed to predict event location and a confidence score for each event query. It employs an MLP to predict offsets~(event center and duration) relative to the event anchor from the query embedding $\mathbf E^L$ output by the decoder. The confidence score indicates whether the predicted event belongs to the foreground.
\label{sec:head}

\noindent \textbf{Captioning Head.}
For the captioning head, we propose an LSTM with deformable soft
attention to encourage the soft attention to focus more on the features around the event anchors. The captioning head takes attention-weighted features $\mathbf z_{n,t}$, query embedding $\mathbf e_n$, and the previous word $w_{n, t-1}$ as input to predict the probability of the next word $w_{n, t}$ based on the hidden state of LSTM. Finally, it generates the entire sentence $\mathcal{S}_n={w_{n,1},...,w_{n,M_j}}$, where $M_j$ represents sentence length.

\noindent \textbf{Event Counter.}
The event counter is proposed to predict the event number of the video. It utilizes a max-pooling layer and a fully-connected layer to produce a high-dimensional vector $\mathbf{v}_{count}$ from query embedding $\mathbf E^L$, each value in $\mathbf{v}_{count}$ represents the possibility of a specific number. The predict event number of the video $N_{select}$ is determined by $N_{select}=\rm{argmax}(\mathbf v_{count})$.

{We denote the loss function corresponding to the prediction heads as $\mathcal{L}_{head}$:
\begin{equation}
    \mathcal{L}_{head} = \mathcal{L}_{cls} + \lambda_{loc} \mathcal{L}_{loc} + \lambda_{cnt} \mathcal{L}_{cnt} + \lambda_{cap} \mathcal{L}_{cap},
\end{equation}
where $\mathcal{L}_{cls}$ and $\mathcal{L}_{loc}$ are the focal loss~\cite{focal} and generalized IOU loss~\cite{giou} respectively, which is calculated in localization head. $\mathcal{L}_{cnt}$ is the cross-entropy loss and is computed in the event counter. $\mathcal{L}_{cap}$ measures the cross-entropy between the predicted word probability and the ground truth captions, which is produced by the captioning head.}

\subsection{Training and Inference}
\label{sec:training}

Our method is trained end-to-end, during training, PR-DETR outputs a set of $N$ event locations and corresponding captions \(\{(t^s_n,t^e_n,\mathcal S_n)\}^{N}_{n=1}\). We adopt the Hungarian algorithm~\cite{hungarian} to match the predicted $N$ events with ground truths, obtaining the best permutation.
Then, we use the best permutation as a new index of predictions to
calculate the total training loss $\mathcal{L}$ as follows:
\begin{equation}
    \mathcal{L} = \mathcal{L}_{head} + \lambda_{prop} \mathcal{L}_{prop},
\end{equation}
where $\mathcal{L}_{prop}$ represents the proposal loss used to supervise the generation of position-anchored queries in Section~\ref{sec:decoding}.

During inference, we select top-$N_{select}$ detected events as final results \(\{(t^s_n,t^e_n,\mathcal S_n)\}^{N_{select}}_{n=1}\) based on the confidence of each query.

\input{table/caption}

\section{Experiment}
\label{sec:experiment}

\subsection{Experimental Settings}

\noindent \textbf{Datasets.}
We evaluate PR-DETR on the two popular dense video captioning datasets: ActivityNet Captions and YouCook2.
ActivityNet Captions contains 20k untrimmed videos of various human activities. On average, each video spans 120s and
is annotated with 3.7 temporally localized sentences.
We follow the standard data split for training, validation, and testing. YouCook2 consists of 2k untrimmed cooking procedure videos, with an average duration of 320s per video and 7.7 temporally localized sentences per annotation. This dataset focuses on cooking scenes, and most videos contain one main actor cooking while describing the recipe. 
Since our method uses CLIP features extracted from raw videos, we only use those videos that are currently accessible on YouTube as training data, which is approximately 7\% less than the the original video count.

\noindent \textbf{Evaluation Metrics.}
We evaluate our method in terms of event localization and caption generation, respectively. For dense captioning performance, we use the official evaluation tool~\cite{wang2020dense} of ActivityNet Challenge, which calculates matched pairs between generated events and the ground truth across IoU thresholds of \{0.3, 0.5, 0.7, 0.9\} and computes captioning metrics~(BLEU4~\cite{vedantam2015cider}, METEOR~\cite{papineni2002bleu}, CIDEr~\cite{banerjee2005meteor}) over the matched pairs.
Moreover, we incorporate SODA\_c~\cite{fujita2020soda}, which considers all event captions from the same video. 
For localization performance, we report the average
precision and average recall across IoU thresholds of \{0.3, 0.5, 0.7, 0.9\}, along with the overall F1 score.

\noindent \textbf{Implementation Details.}
We use pre-trained CLIP Vit-L/14 to extract video features at 1 FPS. We set the number of position-anchored queries to 10 for ActivityNet Captions and 100 for YouCook2, and the iteration $K$ of slot attention is set to 3 in both datasets.
The hidden dimension $D$ of our model is set to 512, and the number of convolution layers $H$ is set to 4. In the relation-enhanced decoder, the dimension of relation embedding is chosen as 16, and the number of attention heads $M$ is set to 8.
Our model is trained end-to-end. For both datasets, we employ the Adam optimizer with an initial learning rate of $5\times10^{-5}$ and 30 training epochs, the batch size is set to 1. The hyper-parameters for weight scaling $\lambda_{loc}$, $\lambda_{cnt}$, $\lambda_{cap}$ and $\lambda_{prop}$ are set to 2, 1, 1, 1 respectively.

\subsection{Comparison with State-of-the-art Methods}

In this part, we compare the dense captioning and event localization performance of PR-DETR with previous methods on YouCook2 and ActivityNet Captions datasets. We only use the visual features of the video as inputs without other modalities such as transcribed speech and audio. In addition, our method is trained solely on the DVC dataset and does not use extra videos for training.

\noindent{\textbf{Dense Captioning Performance.}}
In~\Cref{tab:cap}, we show the dense captioning performance of our method and various dense video captioning approaches on YouCook2 and ActivityNet Captions datasets.
Our method achieves competitive performance on both datasets, outperforming its counterparts PDVC and CM2 in almost all metrics.
By injecting prior knowledge into our model, we reduce the demand for data, achieving comparable performance on less training data compared with Vid2seq and streaming GIT, which use an extra 1 million videos on YouCook2 and 15 million videos on ActivityNet for pertaining.
This is because the proposed event relation encoder emphasizes the relationship between events with co-occurring semantics, maintaining consistency between event captions, thus improving caption quality.
\Cref{tab:cap} shows the performance of some Video Large Language Models~(VLLMs) after fine-tuning on the DVC dataset, which is reported in ~\cite{timechat}. It is worth noting that our method surpasses these VLLMs such as TimeChat by a large margin, which indicates the value of exploring task-specific priors for downstream applications.

\input{table/loc}
\input{table/ab1}

\input{table/iteration_ab}

\noindent{\textbf{Localization Performance.}} We also compare the localization performance of our method with other approaches on both datasets. As shown in~\Cref{tab:loc}, our PR-DETR achieves the best precision and F1 scores and second recall scores in YouCook2 dataset. In ActivityNet Captions, our method achieves the best precision and recall scores, leading to the highest F1 score. 
This is mainly because we initialize scene-specific queries from ground-truth event distribution, which eliminates some implausible event locations and provides a more reliable initial event search region, thus improving the model's localization ability. In contrast, existing VLLMs and pretraining methods focus more on the model's captioning ability, which helps less with event localization.

\input{table/query}



\input{table/loss_ab}
\input{table/module_ab}

\subsection{Ablation Studies}

In this section, we conduct ablation studies to investigate the impact of key components of PR-DETR.

\noindent{\textbf{Effect of Position and Relation Prior.}}
We first conduct ablation experiments to verify the effectiveness of the position prior and relation prior introduced in the position-anchored query and relation-enhanced decoder, respectively. As shown in \Cref{tab:ab1}, the introduction of both position and relation prior can improve all metrics effectively. The position prior improves the localization performance, while the relation prior mainly increases the captioning performance.
This is because the position-anchored query with position prior provides more reliable event search areas, leading to higher precision and recall scores.
The improvement in captioning performance validates the effectiveness of modeling explicit event relations for better caption quality and consistency. Combining the position-anchored query and relation-enhanced decoder, our method achieves significant performance improvement in both localization and captioning sub-tasks.

\noindent{\textbf{Effect of Iteration Number $K$ in Feature Aggregator.}}
We investigate the effect of the iteration number $K$ in the feature aggregator. \Cref{tab:slot} shows the performance corresponding to the varying iteration number $K$.
The iteration number $K$ is related to the granularity of feature aggregation. As $K$ increases, the slot embedding groups frame-wise features in a wider temporal range, which gradually improves the model performance. A large number of $K$ makes the model more likely to ignore shorter events, which leads to performance degradation, especially on YouCook2 dataset with many short events.
Consequently, the iteration number $K$ is determined to be $3$ for the best performance.

\noindent{\textbf{Component-wise Analysis of Position-anchored Query.}}
We conduct component-wise ablative experiments on the YouCook2 dataset to further analyze the effectiveness of each component within the position-anchored query. As shown in \Cref{tab:query}, initializing explicit queries with clustering centroids effectively improves the model performance~(Row $2$). Note that performing feature aggregation without initialized explicit queries only brings a slight performance gain due to the lack of guidance from the clustered event locations~(Row $3$). When we incorporate both explicit query and feature aggregator into our method, the model performance is improved by a considerable margin (Row $5$), which demonstrates the significance of generating scene-specific queries with the position prior. We also take the above scene-specific event anchor as the static anchor for the relation-enhanced decoder to maintain the position prior during decoding. This design further enhances the performance~(compare Row $5$ and Row $6$), leading to the best result.

\noindent{\textbf{Effect of the Proposal Loss.}} 
The balancing parameter $\lambda_\text{prop}$ controls the impact of proposal loss $\mathcal{L}_\text{prop}$, which is designed to supervise the query generation.
To study the effect of $\mathcal{L}_\text{prop}$, we report the performance of different $\lambda_\text{prop}$ in~\Cref{tab:sup_loss}. The result shows that the best performance is obtained when $\lambda_\text{prop} = 1$,  too large or too small $\lambda_\text{prop}$ leads to performance degradation, which validates the impact of the proposal loss. Consequently, we choose $\lambda_\text{prop}$ to be 1 in our paper for the best performance.


\begin{figure}[t]
\centering
\includegraphics[width=\linewidth]{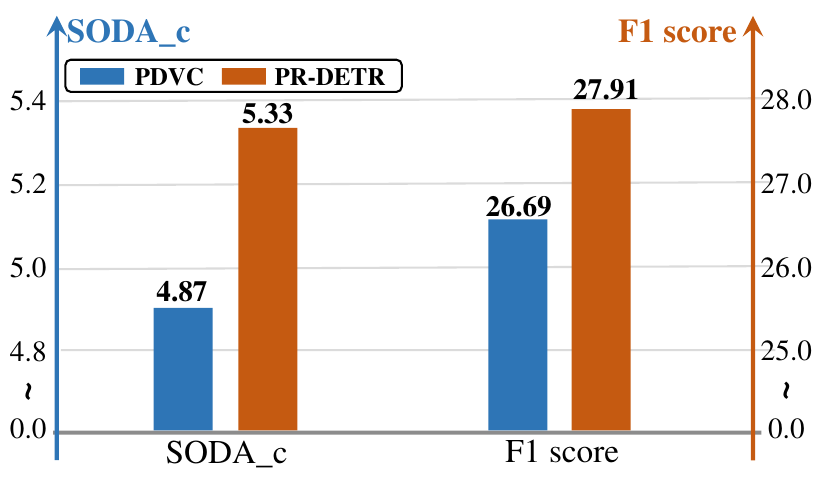}
\caption{Performance comparison between PDVC and our PR-DETR on the ActivityNet-YouCook2 mixed dataset.}
\label{fig:new_res}
\end{figure}

\noindent{\textbf{Comparison of Alternative Relation Encoding Methods.}}
To obtain the {relation mask}, we design two approaches to compute pairwise relative distance between event boundaries, named center-aware distance and overlap-aware distance, respectively. The center-aware distance calculates the relative distance between event centers. 
The overlap-aware distance follows the IOU-like metrics to indicate whether two event boundaries are temporally connected, as described in~\Cref{eq:relation}. The overlap-aware distance of two temporally connected events is 0, whereas overlapping or distant events are computed as close to 1. As shown in~\Cref{tab:module_ab}, the overlap-aware distance achieves better performance compared to the center-aware distance on both captioning and localization metrics. 
This is because overlap-aware distance takes into account the overlap and duration of events, which can more accurately depict the positional relationship between two events compared to center-aware distance. The results demonstrate the value of exploring more accurate metrics of event location relationships.

\noindent{\textbf{Results on the Mixed Dataset.}} We mix ActivityNet and YouCook2 datasets to construct a mixed dataset with more diverse and complex patterns, which is possible in real-world scenarios. Specifically, we sample 10\% of the training videos from ActivityNet and combine them with the training videos from YouCook2 dataset. For the evaluation, we mix the validation sets of the two datasets with the same ratio. Afterwards we apply the training and evaluation settings for ActivityNet to conduct experiments on the mixed dataset. As shown in~\Cref{fig:new_res}, our method achieves substantial improvements over PDVC. By incorporating explicit prior knowledge, our method is more effective in learning complex event patterns of the input data.





\begin{figure*}[t]
\centering
\includegraphics[width=0.96\linewidth]{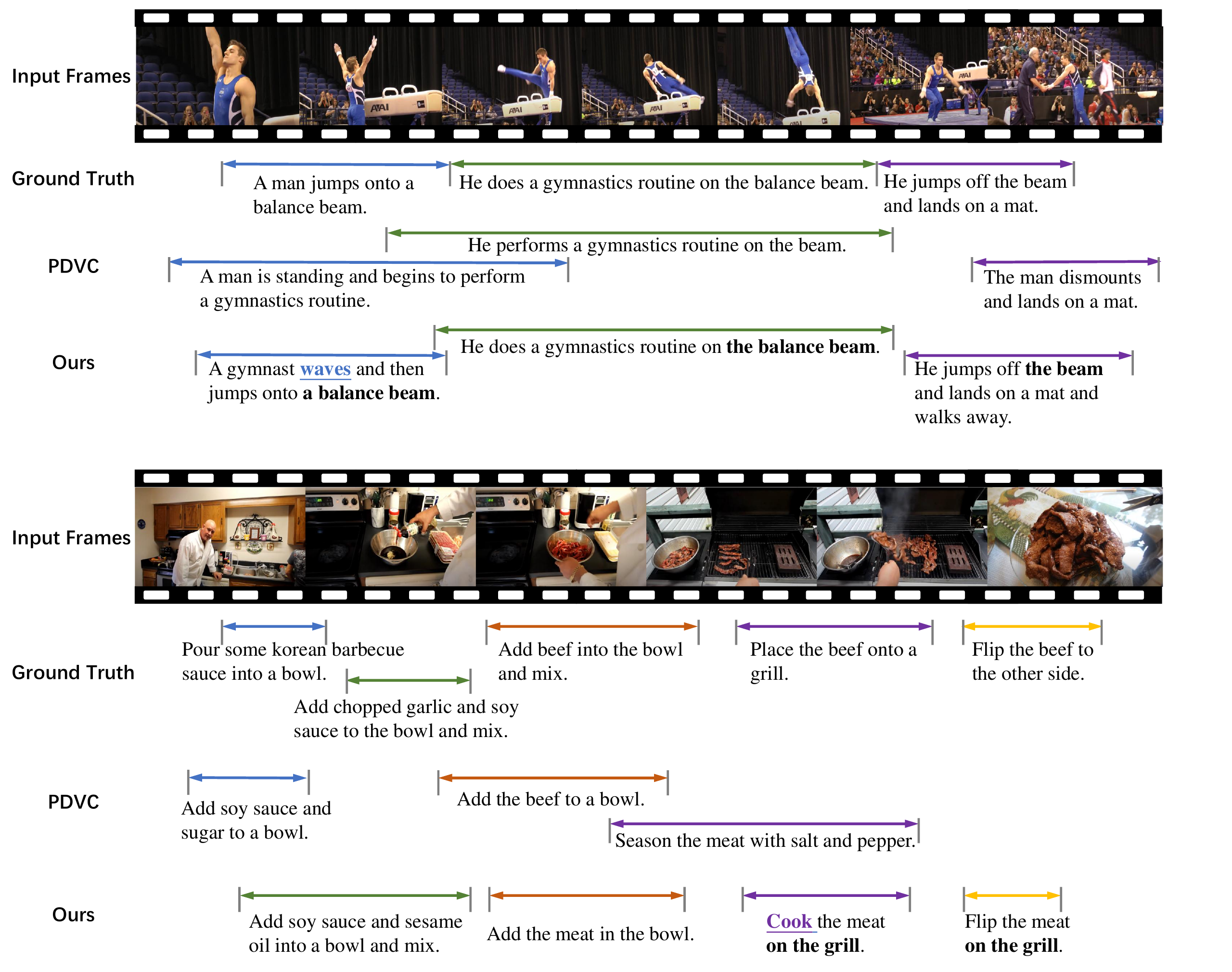}
\caption{Visualization of dense video captioning results of PDVC and our PR-DETR on ActivityNet Captions and YouCook2 validation set. Lines in different colors represent the temporal location of different events. The co-occurring semantics and fine-grained actions recognized by our method are highlighted.}
\label{fig:res_vis}
\end{figure*}

\subsection{Visualization}
We visualize the prediction results of different models in \Cref{fig:res_vis}. Benefiting from the introduction of position prior, our method predicts more accurate event boundaries. Moreover, the explicit relation we construct from event locations helps to model the interaction between events, which improves the coherence of dense captions in the video. 
As a result, the captions generated by our method contain the co-occurring target in the context, \textit{e.g.}, `the balance beam'. Our model also successfully recognizes some fine-grained actions that are not mentioned in the ground-truth, such as `waves' and `cook'.

\section{Conclusion}
\label{sec:conclusion}
In this paper, we propose PR-DETR, which injects the explicit position and relation prior knowledge into the detection transformer to help the model training for better event localization and caption generation. The position-anchored query provide more reliable scene-specific position and semantic information about potential events. The event relation encoder constructs the relation mask between event queries as relation prior to guide the event interaction in the relation-enhanced decoder, improving the semantic coherence of the captions. Our PR-DETR offers an innovative solution for leveraging task-specific knowledge in dense video captioning. Ablation studies validate the effectiveness of position and relation prior. Experiment results on ActivityNet Captions and YouCook2 datasets also demonstrate the improved performance of PR-DETR.

\bibliographystyle{IEEEtran}
\bibliography{reference}

\vfill

\end{document}

%% file: table/caption.tex
\begin{table*}[t!]
\large
\renewcommand{\arraystretch}{1.3}
\caption{{Dense captioning performance comparison with the state-of-the-art methods.}
The best results for each metric are highlighted in bold. {$^{\dagger}$} denotes results reproduced from official implementation using CLIP features.}
\centering
\resizebox{2.0\columnwidth}{!}{
\setlength{\tabcolsep}{4pt}
    \begin{tabular}{l|cc|cccc|cccc}
    \toprule
   \multirow{2}{*}{Method}&\multirow{2}{*}{Features} &\multirow{2}{*}{Pretrain} & \multicolumn{4}{c}{YouCook2} & \multicolumn{4}{|c}{ActivityNet} \\
    &&&BLEU4$\uparrow$ & METEOR$\uparrow$ & CIDEr$\uparrow$ & SODA$\_c$$\uparrow$ &BLEU4$\uparrow$ & METEOR$\uparrow$ & CIDEr$\uparrow$ & SODA$\_c$$\uparrow$\\
    \hline
    {\textbf{\textit{Video LLMs}:}}&&& & & &  & & &  &\\
    Video-LLaMA~\cite{videollama}  &-&-&- & - & 6.10 & 1.80 &- &- & - & - \\ 
    VideoChat-Embed~\cite{videochat}  &-&-&- & - & 8.10 & 2.20 &- &- & - & -\\
    TimeChat~\cite{timechat}  &-&-&- & - & 10.30 & 3.10 &- &- & - & -\\
    
    \hline
    {\textbf{\textit{Pre-training}:}} &&& & & &  & & &  & \\
    Streaming GIT~\cite{zhou2024streaming} &CLIP &\Checkmark  & - & 3.60 & 15.40  &3.20 & - & 9.00 & 41.20 & 6.60  \\
    
    Vid2Seq~\cite{yang2023vid2seq} &CLIP  &\Checkmark  & - & 9.30 &47.10 & 7.90 & - &  8.50 & 30.10 & 5.80  \\
    \hline
    
    E2ESG~\cite{E2ESG} &C3D  &\XSolidBrush &- &3.50 &25.00  &-  &- &- &- &-  \\
    ECHR~\cite{ECHR} &C3D   &\XSolidBrush &- &3.82 &- &- &1.29 &7.19 &14.71 &3.22\\
    MT~\cite{MT} &TSN &\XSolidBrush &0.30 &3.18 &6.10 &- &1.15 &4.98 &9.25 &-\\
    PDVC~\cite{pdvc} &TSN &\XSolidBrush &0.80 &4.74 &22.71 &4.42  &1.78 &7.96 &28.96 &5.44\\
    
    PDVC{$^{\dagger}$}~\cite{pdvc} &CLIP  & \XSolidBrush &1.31 &5.47 &29.20 &4.95 &2.11  &8.09 &29.68 &5.87\\
    
    {CM2}~\cite{cm2}&CLIP &  \XSolidBrush & 1.63 & 6.08 & 31.66  & 5.34 & 2.38 & 8.55 &33.01 &\textbf{6.18} \\
    
    \textbf{PR-DETR~(Ours)}&CLIP  &\XSolidBrush &\textbf{1.89}& \textbf{6.48}  &\textbf{37.30}  &\textbf{5.43} & \textbf{2.58}& \textbf{8.72} &\textbf{33.16}   &6.13 \\
    \bottomrule
    \end{tabular}
}
\label{tab:cap}
\end{table*}

        
        


%% file: table/loc.tex
\begin{table*}[t]
\large
\renewcommand{\arraystretch}{1.3}
\caption{{Event localization performance comparison with the state-of-the-art methods.}
The best results for each metric are highlighted in bold. {$^{\dagger}$} denotes results reproduced from official implementation using CLIP features.}
\centering

\resizebox{1.5\columnwidth}{!}{
\setlength{\tabcolsep}{4pt}
    \begin{tabular}{l|c|ccc|ccc}
    \toprule
    \multirow{2}{*}{Method}&\multirow{2}{*}{Pretrain} & \multicolumn{3}{c}{YouCook2} &\multicolumn{3}{|c}{ActivityNet}  \\
    & & Precision$\uparrow$ & Recall$\uparrow$ &F1 score$\uparrow$  & Precision$\uparrow$ &Recall$\uparrow$ & F1 score$\uparrow$\\
    \hline
    {\textbf{\textit{Video LLMs}:}}&&& & & &&\\
    Video-LLaMA~\cite{videollama} &- & - & - & 14.20  & - & - & -\\
    TimeChat~\cite{timechat} &- & - & - & 19.50  & - & - & -\\
    \hline
    {\textbf{\textit{Pre-training}:}}&&& & & &&\\
    Streaming GIT~\cite{zhou2024streaming}&\Checkmark  & - & - & 16.60  & - & - & 50.90\\
    Vid2Seq~\cite{yang2023vid2seq}&\Checkmark  & 27.80 & \textbf{27.90} & 27.30  & 53.90&  52.70 & 52.40\\
    \hline
    E2ESG~\cite{E2ESG} &\XSolidBrush  & 20.60 & 20.70 & 20.65  & - & - & - \\
    
    MFT~\cite{xiong2018move} &\XSolidBrush  & - & - & -  & 51.41 & 24.31 & 33.01 \\
    
    PDVC{$^{\dagger}$}~\cite{pdvc} &\XSolidBrush    & 32.58 & 23.48 & 27.29   & {56.72} & {53.22} 
    & {54.91}\\
    
    {CM2}~\cite{cm2}&\XSolidBrush   & {33.38} & {24.76}& {28.43}   
    &{56.81} & {53.71} & {55.21}\\
    
    \textbf{PR-DETR(Ours)}&\XSolidBrush &\textbf{36.11} &24.06 &\textbf{28.87}  &\textbf{57.48} &\textbf{53.82} &\textbf{55.59}

 \\
    \bottomrule
    \end{tabular}
}
\label{tab:loc}
\end{table*}

%% file: table/ab1.tex
\begin{table}[t!]

\renewcommand{\arraystretch}{1.3}
\caption{Ablation studies on position and relation prior of PR-DETR on YouCook2 validation set. {P} denotes the position-anchored query and {R} denotes the relation-enhanced decoder.}
\centering
\resizebox{1\columnwidth}{!}{
\setlength{\tabcolsep}{3pt}
    \begin{tabular}{l|cccccc}
    \toprule
    Method &B4$\uparrow$& M$\uparrow$ & C$\uparrow$  &S$\uparrow$ &Pre.$\uparrow$ &Rec.$\uparrow$ \\
    \midrule
    
     Base &1.31 & 5.47 & 29.20 &4.95
     &32.58&23.48 \\
     
      Base+\textbf{P}& 1.39& 5.95& 31.54& 5.40&34.34 & \textbf{24.36}\\
     
   Base+\textbf{R}& 1.68& 6.13& 33.44& 5.16 &34.76 & 22.94
\\
    
     Base+\textbf{P}+\textbf{R}~(Ours)& \textbf{1.89}& \textbf{6.48}& \textbf{37.30}& \textbf{5.43} &\textbf{36.11}& 24.06
\\
    
    \bottomrule
    \end{tabular}
}
\vspace{-0.2cm}
\label{tab:ab1}
\end{table}

%% file: table/iteration_ab.tex
\begin{table}[t!]

\renewcommand{\arraystretch}{1.3} \caption{Effect of the iteration number $K$ in feature aggregator on YouCook2 validation set.}
\centering
\resizebox{1\columnwidth}{!}{
\begin{tabular}{c|cccccc}
\toprule
\multirow{1}{*}{Iteration $K$} &B4$\uparrow$& M$\uparrow$ & C$\uparrow$  & S$\uparrow$ &Pre.$\uparrow$ &Rec.$\uparrow$ \\
\midrule
1 &1.79 &\textbf{6.69}  &36.29 &5.14 &\textbf{37.36} &22.23 \\
2 &1.78 &6.41  &36.09 &5.41 &35.89  &\textbf{24.47}  \\
3 &\textbf{1.89}& 6.48& \textbf{37.30}& \textbf{5.43}&36.11 & 24.06 \\
4 &1.77 &6.64  &35.56 &5.29 &36.86 &22.81  \\

\bottomrule
\end{tabular}
}
\vspace{-0.2cm}
\label{tab:slot}
\end{table}

%% file: table/query.tex
\begin{table}[t!]
\large
\renewcommand{\arraystretch}{1.3}
\caption{Component-wise ablation results of the position-anchored query on YouCook2 validation set. EQ means initializing explicit queries with the clustering centers, FA means feature aggregator and SA denotes static anchor $\mathbf P_\text{sta}$.}
\centering
\resizebox{1.0\columnwidth}{!}{
\begin{tabular}{ccc|cccccc}
\toprule
 \makecell {EQ} & \makecell{FA} & \makecell{SA}  &B4$\uparrow$& M$\uparrow$ & C$\uparrow$  &  S$\uparrow$ &Pre.$\uparrow$ &Rec.$\uparrow$ \\
\midrule
    && &1.68 & 6.13 & 33.44 &5.16 &34.76 &22.94 
 \\
 \checkmark&&&1.75  &6.23  &34.40 &5.25 &34.76 &22.81  \\
 
 &\checkmark&& 1.74 & 6.16 & 35.03 & 5.25 &34.78 &23.02 
 \\

\checkmark&&\checkmark &1.67 & 6.16 & 34.86 &5.36 &35.04 &23.56 \\

\checkmark&\checkmark&& 1.84& 6.45 & 36.72 &5.36 &\textbf{36.19} &23.83   \\

\checkmark& \checkmark&\checkmark& \textbf{1.89}& \textbf{6.48}& \textbf{37.30}& \textbf{5.43}& 36.11 & \textbf{24.06}  \\
\bottomrule
\end{tabular}
}
\vspace{-0.2cm}
\label{tab:query}
\end{table}

%% file: table/loss_ab.tex


\begin{table}[t!]

\renewcommand{\arraystretch}{1.3}
\caption{{Performance of different balancing parameter $\lambda_\text{prop}$ on YouCook2 validation set.} $\lambda_\text{prop}$ controls the impact of proposal loss $\mathcal{L}_\text{prop}$. 
}
\centering
\resizebox{1\columnwidth}{!}{
\begin{tabular}{c|cccccc}
\toprule
\multirow{1}{*}{$\lambda_{prop}$} &B4$\uparrow$ & M$\uparrow$& C$\uparrow$  &  S$\uparrow$ &Pre.$\uparrow$ &Rec.$\uparrow$ \\
\midrule
0.5 & 1.75 & 6.41 & 35.47 & 5.10 &34.82 &22.31 \\
1 & \textbf{1.89}& 6.48& \textbf{37.30}& \textbf{5.43}&{36.11}& \textbf{24.06}\\
2 & 1.77 & \textbf{6.59} & 36.18 & 5.14 & \textbf{{37.13}} &22.67 \\
3 & 1.78 & 6.56 & 36.12 & 5.23 & 36.78 &23.18 \\
\bottomrule

\end{tabular}
}
\vspace{-0.2cm}
\label{tab:sup_loss}
\end{table}

%% file: table/module_ab.tex



\begin{table*}[t!]

\renewcommand{\arraystretch}{1.2}
\caption{{Performance of different relation encoding methods in the event relation encoder on the YouCook2 validation set.} ‘Center-aware’ means calculating the relative distance between event centers, ‘Overlap-aware’ denotes our proposed metric which indicates whether two events are temporally connected.}
\centering
\resizebox{1.3\columnwidth}{!}{
\begin{tabular}{c|cccccc}
\toprule
Methods &B4$\uparrow$& M$\uparrow$ & C$\uparrow$  & S$\uparrow$ &Pre.$\uparrow$ &Rec.$\uparrow$ \\
\midrule
Center-aware Distance & 1.72 & 6.13 & 35.25 & 5.30  &34.23&22.28 \\

Overlap-aware Distance&{1.89}& {6.48}& {37.30}& {5.43}&{36.11}& 24.06\\
\bottomrule

\end{tabular}
}
\vspace{-0.2cm}
\label{tab:module_ab}
\end{table*}